\documentclass[runningheads]{llncs}

\usepackage{eccv}

\usepackage{eccvabbrv}
\usepackage{footmisc}
\usepackage{graphicx}
\usepackage{booktabs}
\usepackage{makecell}
\usepackage{multicol}
\usepackage{multirow}
\usepackage{eurosym}
\usepackage{comment}
\usepackage{etoolbox}
\usepackage{bbding,pifont}
\usepackage{tabularx}
\newcommand{\cmark}{\ding{51}}%
\newcommand{\xmark}{\ding{55}}%
\usepackage[accsupp]{axessibility}  

\usepackage[breaklinks,colorlinks,citecolor=eccvblue]{hyperref}

\usepackage{orcidlink}
\makeatletter
\def\blfootnote{\gdef\@thefnmark{}\@footnotetext}
\makeatother

\begin{document}

\title{Delving Deep into Engagement Prediction of Short Videos} 

\titlerunning{Engagement Prediction of Short Videos}

\author{
Dasong Li$^{1,}$\thanks{First author. Main work was completed during an internship at Snap.} \quad
Wenjie Li$^{2}$\quad
Baili Lu$^{2}$\quad
Hongsheng Li$^{1,3}$\quad
Sizhuo Ma$^{2}$\quad
Gurunandan Krishnan$^{2}$\quad
Jian Wang$^{2,}$\thanks{Corresponding author}\\
$^1$MMLab, CUHK\quad
$^2$Snap Inc.\\
$^3$Centre for Perceptual and Interactive Intelligence Limited
\\
{\tt\small dasongli@link.cuhk.edu.hk, jwang4@snapchat.com}
}
\authorrunning{Li et al.}
\institute{\url{https://github.com/dasongli1/SnapUGC_Engagement}}

\maketitle

\begin{abstract}
  Understanding and modeling the popularity of User Generated Content (UGC) short videos on social media platforms presents a critical challenge with broad implications for content creators and recommendation systems. This study delves deep into the intricacies of predicting engagement for newly published videos with limited user interactions. Surprisingly, our findings reveal that Mean Opinion Scores from previous video quality assessment datasets do not strongly correlate with video engagement levels.
To address this, we introduce a substantial dataset comprising 90,000 real-world UGC short videos from Snapchat. 
Rather than relying on view count, average watch time, or rate of likes, we propose two metrics: normalized average watch percentage (NAWP) and engagement continuation rate (ECR) to describe the engagement levels of short videos.
Comprehensive multi-modal features, including visual content, background music, and text data, are investigated to enhance engagement prediction. With the proposed dataset and two key metrics, our method demonstrates its ability to predict engagements of short videos purely from video content.
\end{abstract}

\section{Introduction}
\label{sec:intro}
With the rapid advancement of social media, an increasing number of content creators post short videos to document and share their daily lives on streaming media platforms such as TikTok, Instagram Reels, Youtube Shorts, and Snapchat Spotlight. Simultaneously, a substantial portion of users spend a significant amount of time in consuming short videos across these platforms.

Social media platforms receive a constant stream of newly published short videos.
Therefore, it is important to determine to what extent each video should be recommended to users.
Recommending high-quality User Generated Content (UGC) videos enhances viewer engagement and consequently encourages content creators, especially novice creators. The effective dissemination of newly published videos remains a core goal of social media platforms.
However, owing to their limited user reactions, accurate recommendation of such \emph{cold-start items} is usually a challenge. Typically, platforms would present each new video to a restricted number of users, \eg one hundred. 
The latent popularity of each video is estimated based on the engagement metrics such as watch times from these initial users, serving as a basis for further recommendations. The cold start problem \cite{cold_start1,cold_start2,cold_start3,cold_start4} arises from the sampling bias in such limited initial interactions, resulting in noisy and inaccurate predictions of recommendation extents. This creates a negative feedback loop within the ecosystem, hindering the recommendation of high-quality videos to users.
Content creators may also face delays in gauging their videos' popularity, slowing their adjustments based on viewer feedback and thus discouraging them from posting more quality content.

\begin{table}[t]
  \centering
  \scriptsize
  \setlength{\tabcolsep}{8pt}
  \begin{tabular}{cc|cccc}
    \toprule
    \multirow{2}{*}{Method} & Trained & \multicolumn{4}{c}{Correlation of different durations} \\ 
    & Dataset & [19, 21) & [29, 31) & [39, 41) & [49, 51) \\ \midrule
    UVQ \cite{uvq} & UGC \cite{uvq} &  0.084 & 0.156 & 0.290 &  0.289 \\
    DOVER \cite{Wu_2023_ICCV} & LSVQ \cite{patch-vq} & 0.073 & 0.148 & 0.305 & 0.286 \\
    \bottomrule
  \end{tabular}
  \caption{Correlation between the predicted mean opinion score (MOS) scores and average watch time. The correlations are separately calculated for videos from 4 disjoint ranges of durations. ``[19, 21)'' refers to the videos of durations in the range of 19s to 21s, and similarly for ``[29, 31)'', ``[39, 41)'', and ``[49, 51)''. Small ranges are chosen to minimize the variation within each group.}
  \label{tab:pilot_study}
\end{table}

Previous video quality assessment (VQA) datasets \cite{kv1k,ytugc,vqc, uvq, patch-vq} rely on subjective scores from relatively small groups of annotators (\eg 40). These subjective scores often exhibit biases due to raters' diverse preferences and limited participation, which may not faithfully reflect a video's popularity among its true audience, gauged via metrics like average watch times. 
Our experiments in Table~\ref{tab:pilot_study} reveal that VQA models \cite{uvq,patch-vq,Wu_2023_ICCV} trained on these existing datasets yield very poor correlation with the popularity of short videos. While these VQA methods mainly focus on video visuals, short video engagement can be influenced by other factors like background music, content category, title, \etc. 
Existing engagement prediction datasets such as Wu \etal~\cite{Wu_2018_beyond,lecture_dataset} focus on limited categories of longer videos, which is not suitable for studying the engagement of short-form videos  across diverse categories.
Moreover, certain prerequisites \cite{Wu_2018_beyond} for historical creator information limits their applicability to videos from new creators.

To overcome the issues encountered in previous VQA datasets, we collect a large-scale UGC short video dataset named SnapUGC, which comprises publicly accessible short videos from Snapchat Spotlight. To mitigate potential biases arising from limited number of  annotators, we propose to leverage engagement data from \emph{real users}. For quantifying engagement levels of short videos, we propose to employ two key metrics: normalized average watch percentage (NAWP) and engagement continuation rate (ECR). NAWP provides an indication of the overall engagement level for videos with different durations. Meanwhile, ECR represents the probability of watch time exceeding 5 seconds, which assesses whether the video's outset is captivating enough to retain viewers' interest in continuing to watch. 
It is worth noting that the two metrics are derived through aggregation from more than 2000 viewers and the dataset does not contain individual viewers' history or personal information, ensuring user privacy.

To predict engagement levels with limited user interactions, we formulate the challenge as extracting engagement solely from video content, independent of user, creator, or contextual cues. To enhance the modeling of engagement in short videos, we move beyond previous visual features \cite{uvq, patch-vq, MD-VQA, resnet3d, MD-VQA}. Our methodology incorporates comprehensive multi-modal features such as video captioning, sound classification, titles, descriptions, and more to model the engagement levels of short videos. The seamless integration of these multi-modal features is achieved through the adoption of a cross-modal attention mechanism, enabling the harmonious fusion of visual and language-based attributes. In contrast to previous Video Quality Assessment (VQA) methods, our approach capitalizes on the incorporation of these comprehensive multi-modal features, resulting in superior performance in the engagement prediction for short videos.

The contributions of this study include: 
1) We introduce a large-scale dataset to facilitate research in predicting engagement for real UGC short videos.
2) We employ two novel metrics, normalized average watch percentage and engagement continuation rate, to characterize engagement levels of short videos.
3) We investigate a diverse set of multi-modal features to strengthen the capacity of engagement prediction.
4) Using the proposed dataset and engagement metrics, our method demonstrates the ability to estimate short videos engagement in a cold start setup, highlighting its significance in the field.

\section{Related Works}
\noindent\textbf{Video quality assessment methods} 
Classical VQA methods \cite{vbliinds,viideo,tlvqm,videval,rapique,tpqi} utilize handcrafted features to evaluate video quality. Given the subjectivity and complexity of video quality, handcrafted features fall short in capturing the nuances of video quality assessment. Most previous deep VQA methods \cite{patch-vq,mlsp,vsfa,MD-VQA,uvq,additional_cite1,additional_cite2,additional_cite3} follow a two-step process: they begin by extracting deep features and subsequently train a temporal regression network using these fixed features. These deep features involves per-frame semantic features \cite{uvq,patch-vq,MD-VQA} from image classification networks \cite{he2016residual,efficientnet_v2} trained on ImageNet-1k \cite{imagenet}, per-frame low-level distortion features \cite{uvq} from low-level distortion recognition networks, and multi-frame semantic features \cite{MD-VQA} from action recognition networks \cite{resnet3d} trained on Kinetics-400 \cite{k400data}.
Gated Recurrent unit (GRU) \cite{gru}, InceptionTime \cite{IsmailFawaz2020inceptionTime} and simple average operations \cite{MD-VQA} are utilized for temporal regression of Mean Opinion Scores (MOS). Recent approaches \cite{wu2022fastquality,wu2022fasterquality,Wu_2023_ICCV} have emerged that opt for an end-to-end methodology, jointly optimizing feature extraction and final regression. However, these aforementioned VQA methods focus on exploring visual features while disregarding the potential contributions of additional information provided by content creators, such as background sound, title, descriptions, \etc. The underexplored domain of vision-language correspondence \cite{Zhang_2023_CVPR,clip} in video quality assessment becomes apparent.
\begin{figure}[!t]
  \centering
    \small
    \begin{center}
    \setlength{\tabcolsep}{1.5pt}
    \begin{tabular}{@{} c c c c c c @{}}
    \includegraphics[width=.16\linewidth]{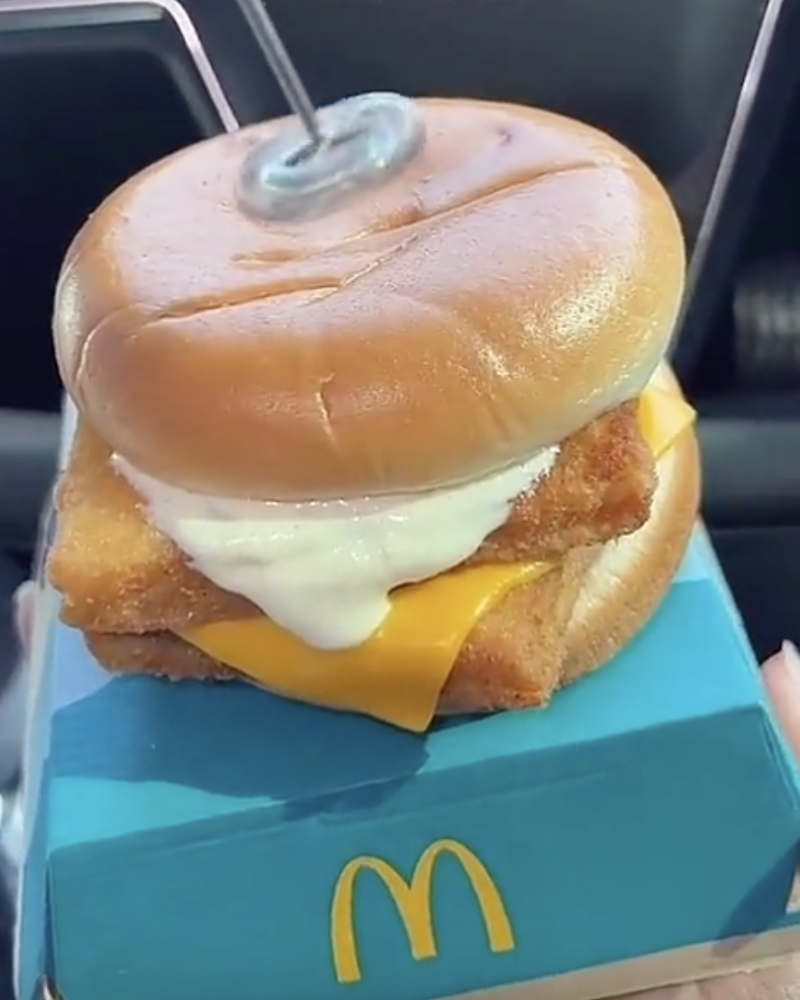} &
    \includegraphics[width=.16\linewidth]{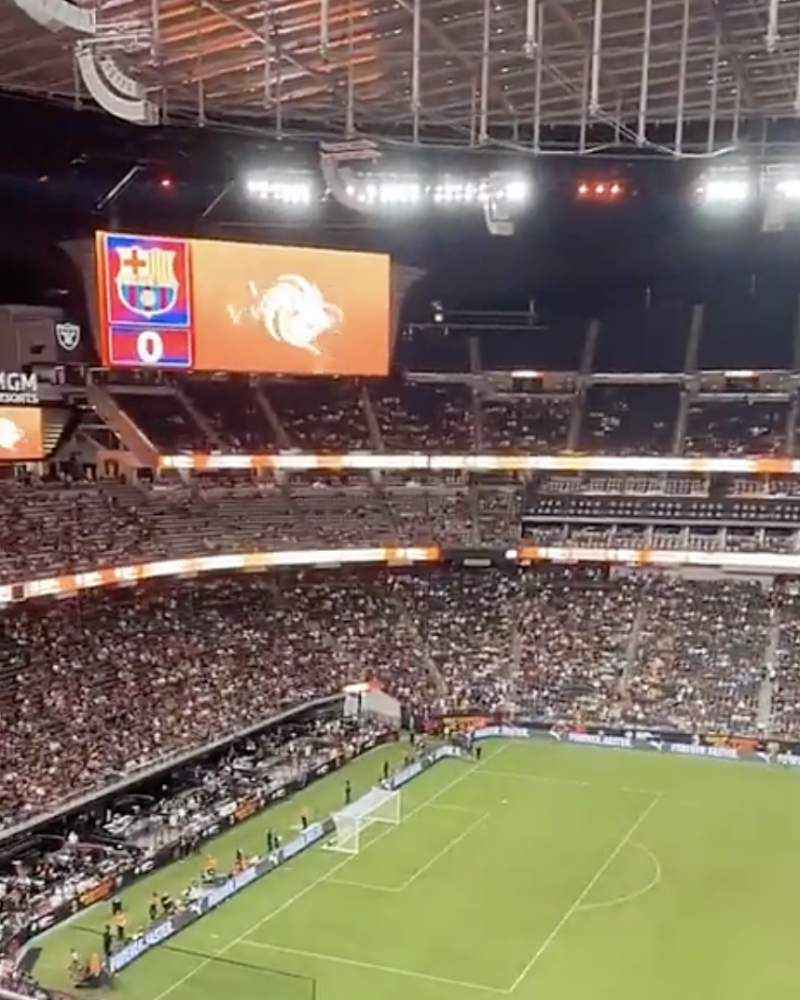} &
    \includegraphics[width=.16\linewidth]{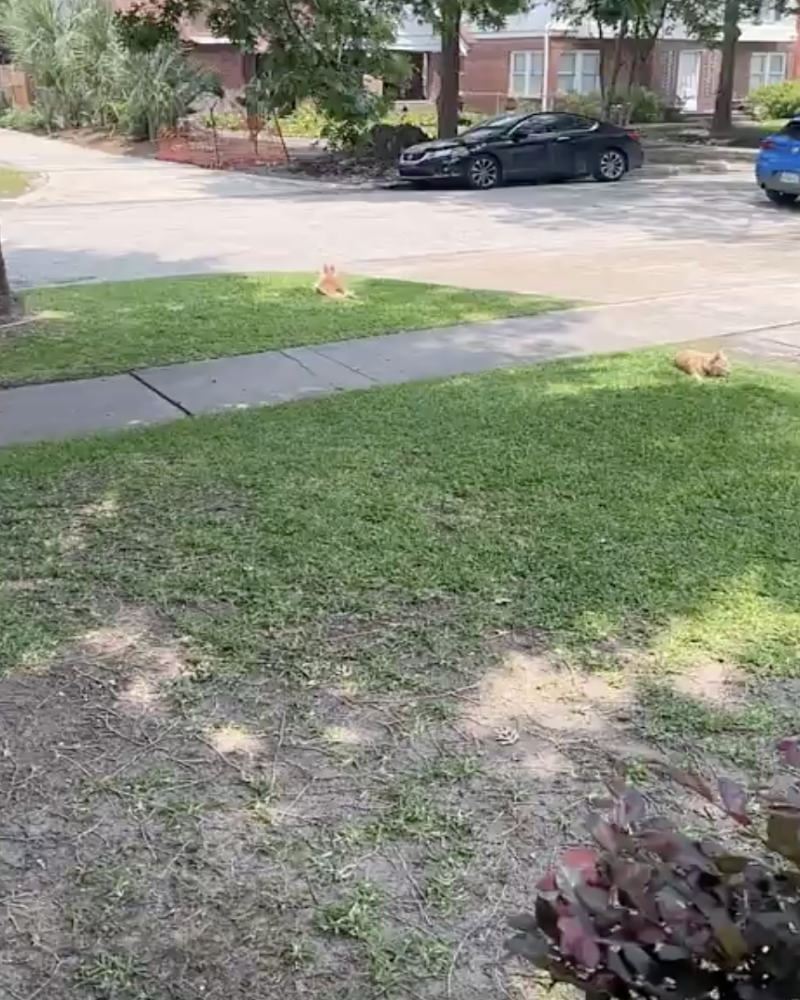} &
    \includegraphics[width=.16\linewidth]{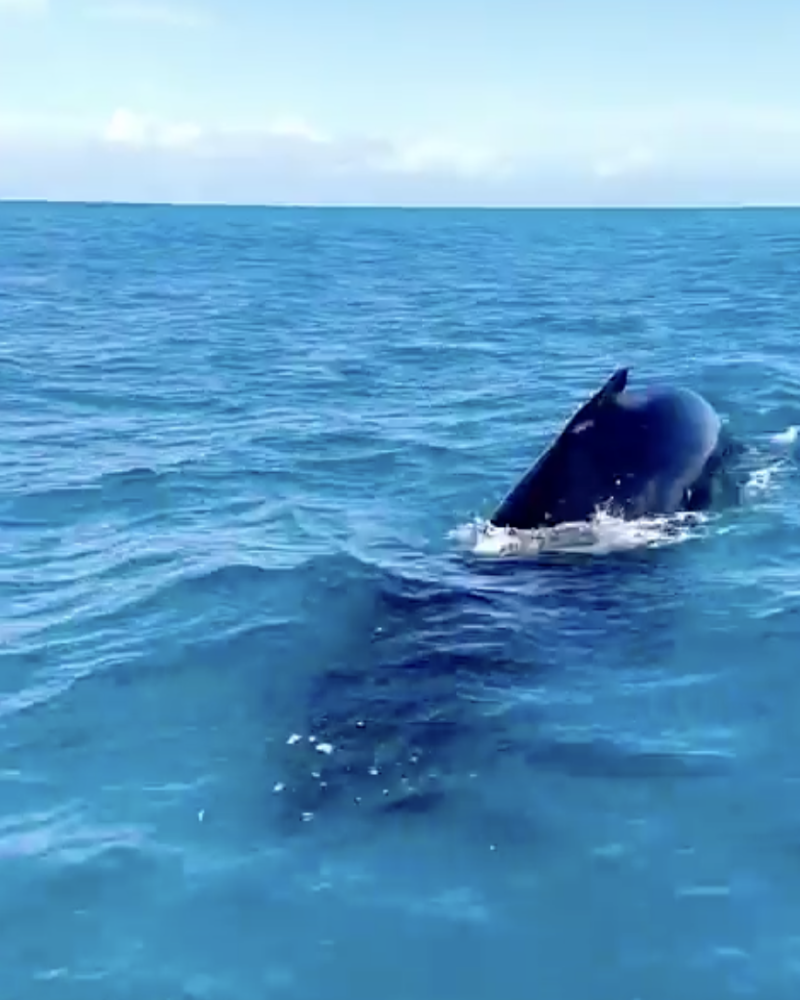} &
    \includegraphics[width=.16\linewidth]{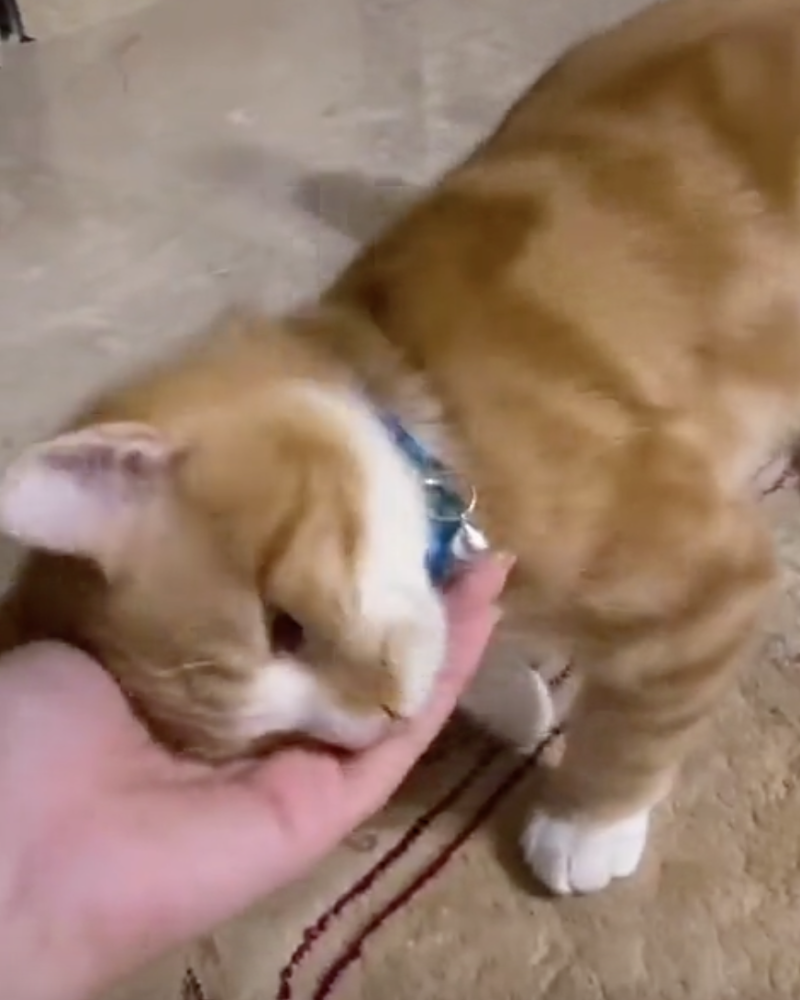} &
    \includegraphics[width=.16\linewidth]{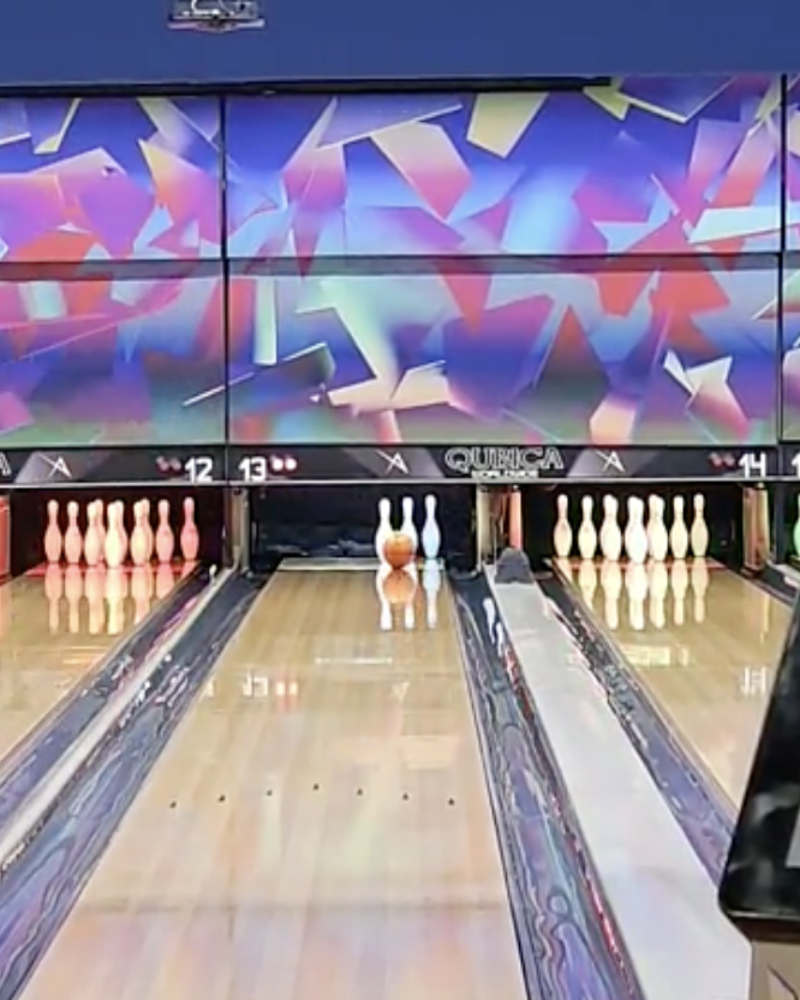} \\
    \includegraphics[width=.16\linewidth]{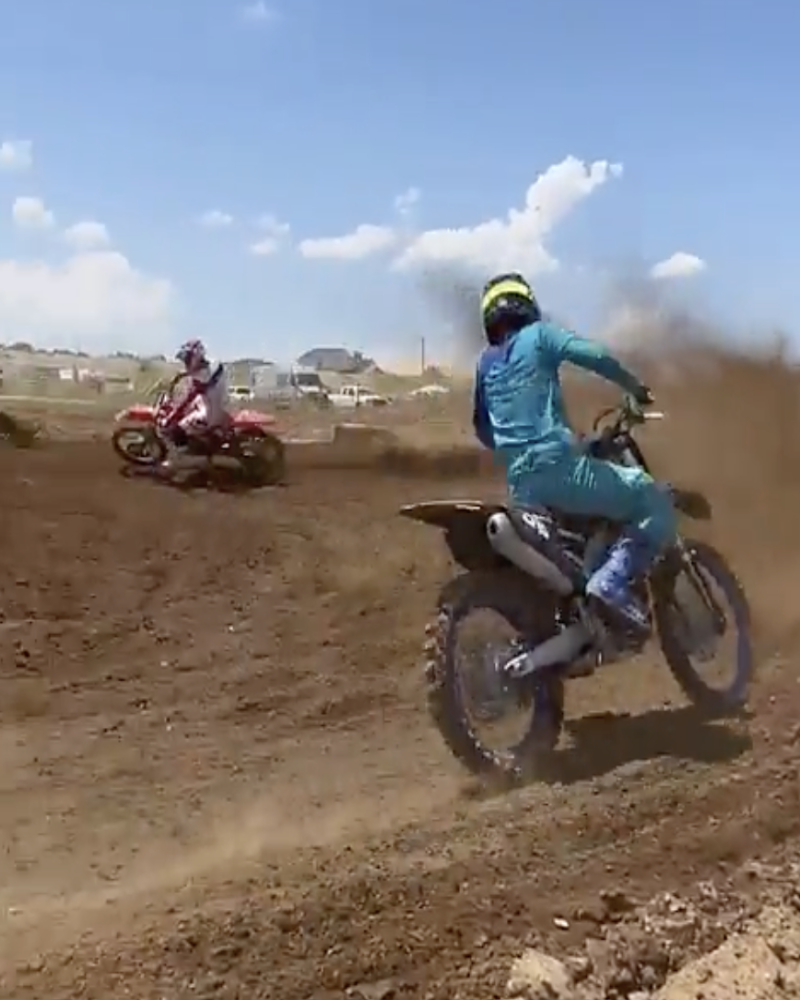} &
    \includegraphics[width=.16\linewidth]{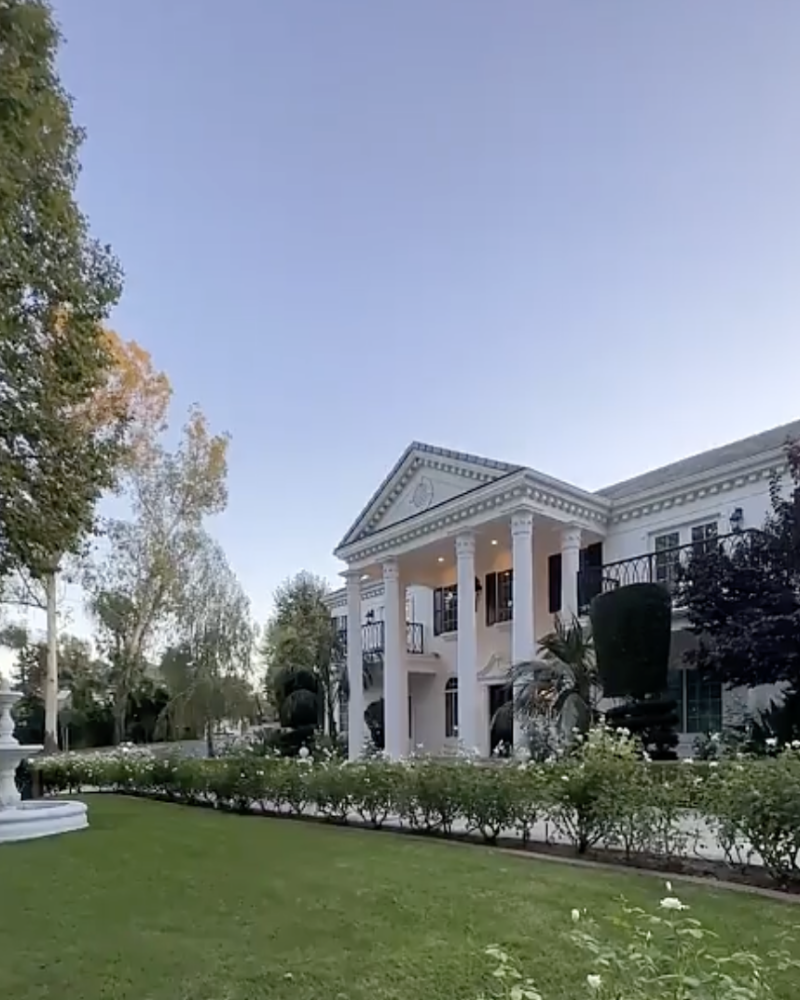} &
    \includegraphics[width=.16\linewidth]{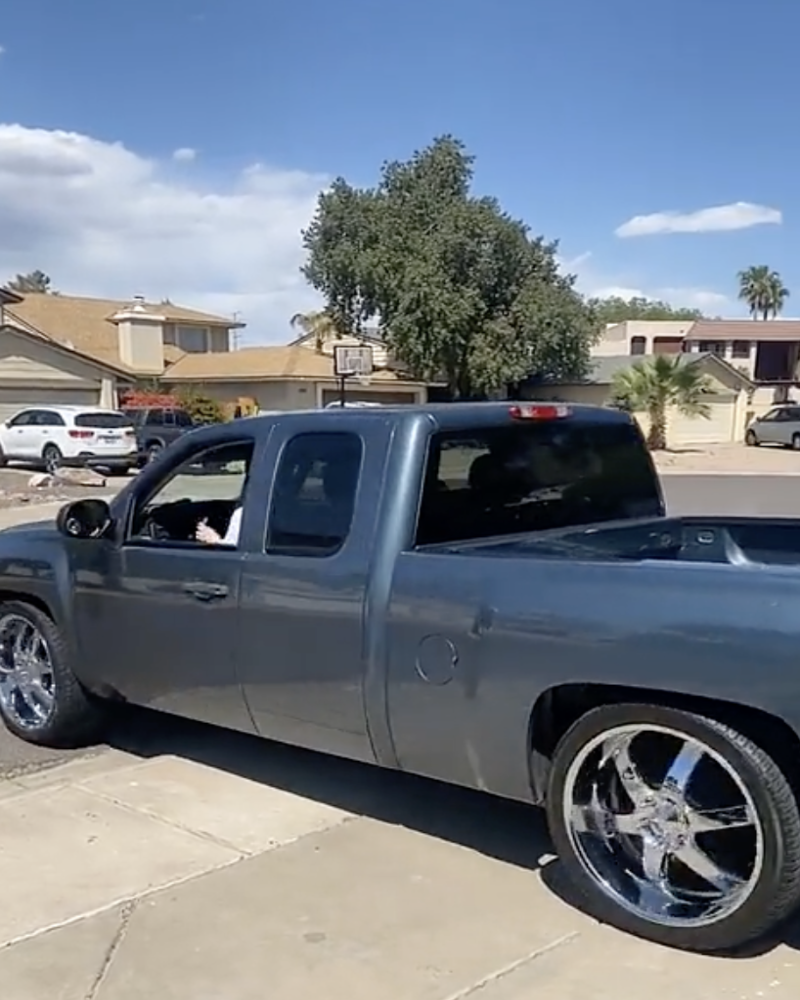} &
    \includegraphics[width=.16\linewidth]{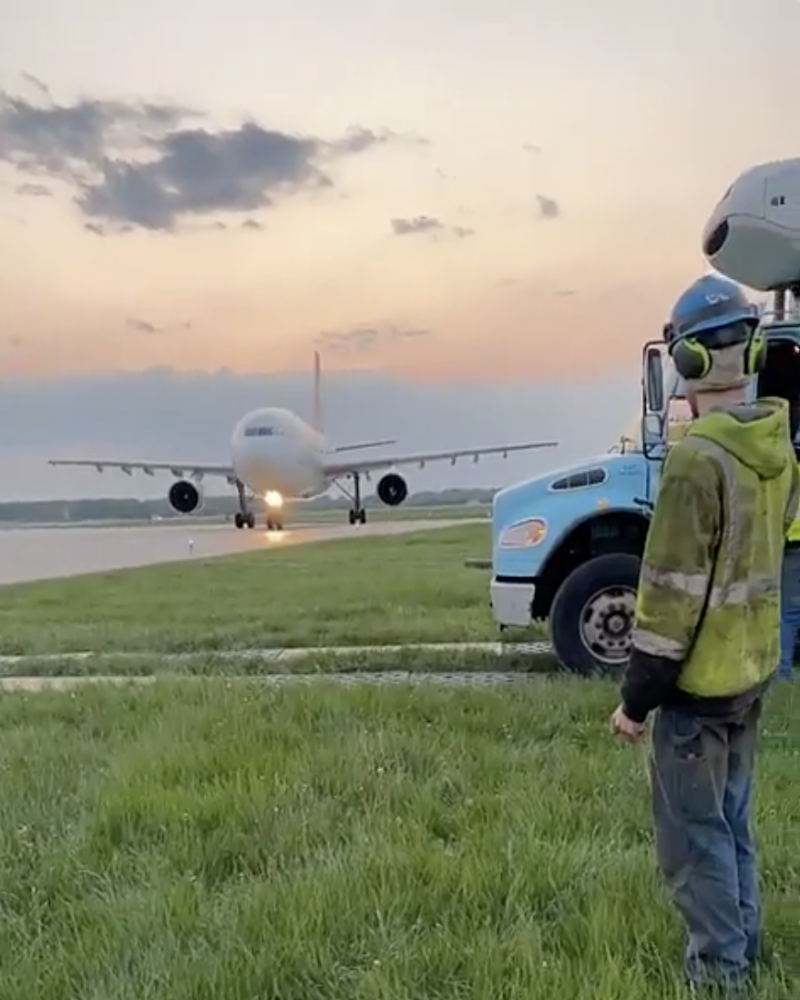} &
    \includegraphics[width=.16\linewidth]{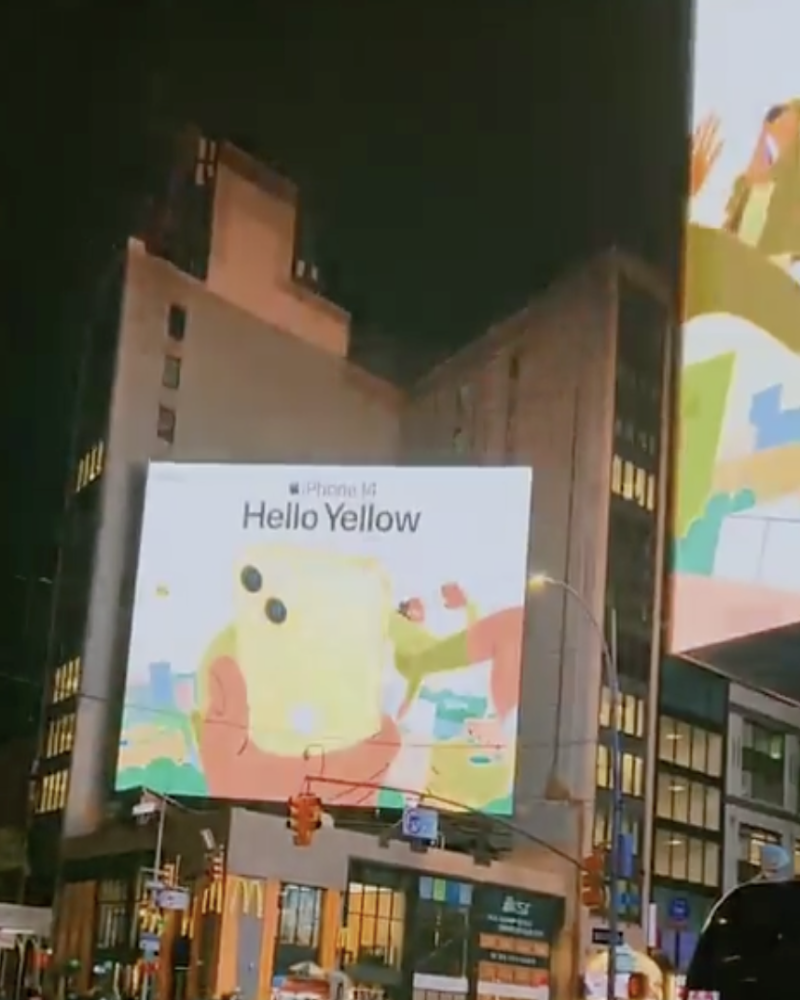} &
    \includegraphics[width=.16\linewidth]{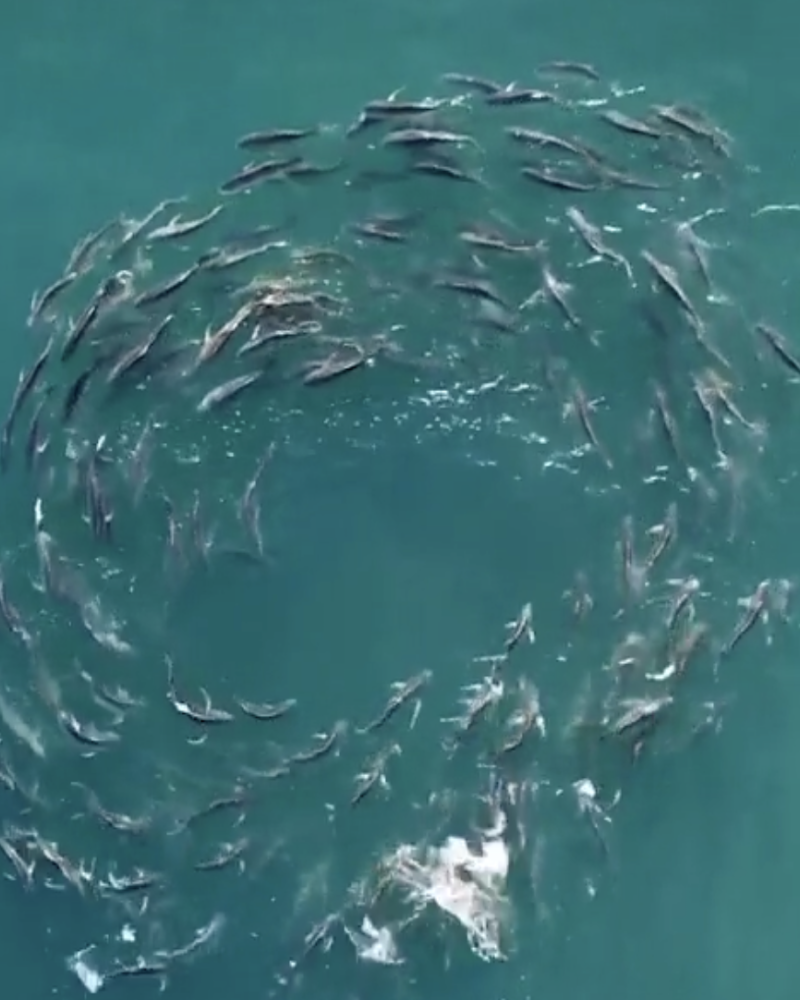} \\
    \end{tabular}
    \end{center}
    \caption{Sample frames of the short videos in our dataset. The frame samples are cropped to exclude sensitive content such as human faces and watermarks for display.}
    \label{fig:sample}
\end{figure}

\noindent\textbf{Video quality datasets} Early datasets \cite{cvd,qualcomm} are often designed with specialized distortions (such as noise \cite{Li2022efficient,Zhang2023kbnet,zhang2021IDR,zhang2023UCDIR} and blur\cite{Li2022Learning,Li_2023_CVPR}) to facilitate the examination of low-level video consistency \cite{shi2023flowformer++,shi2023videoflow,shi2024motion} and quality. In contrast, more recent VQA datasets, such as KoNViD-1k \cite{kv1k}, YouTube-UGC \cite{ytugc}, LIVE-VQC \cite{vqc}, YT-UGC$^+$ \cite{uvq}, and LSVQ \cite{patch-vq}, are introduced with the aim of characterizing the subjective quality of videos. These datasets typically involve the labeling of Mean Opinion Scores (MOS) by a relatively small group of individuals. However, a notable domain gap exists between short videos and the videos in these VQA datasets. 
On social media platforms, users may swiftly skip uninteresting videos instead of watching the whole video, while  annotators of VQA datasets tend to watch the entire video. This unique property of short videos introduces a discrepancy between engagement levels and previous MOS scores.

\noindent\textbf{Engagement prediction} 
Previous datasets focus on analyzing engagement of video lectures \cite{lecture_dataset} and YouTube videos \cite{Wu_2018_beyond}. Regrettably, there is a scarcity of publicly available datasets specifically tailored for predicting engagement for short videos. Commonly employed metrics for video engagement include view counts, average watch time, and average watch percentage. Video duration emerges as a critical covariate affecting both average watch time and average watch percentage, as illustrated in \etal \cite{Wu_2018_beyond,kuaishou}. Intuitively, longer videos are less likely to be watched in their entirety compared to shorter videos, a phenomenon attributed to the diminishing attention span of viewers. In response, Wu \etal \cite{Wu_2018_beyond} propose a relative engagement metric that accounts for varying video durations. 
However, the relative engagement metric takes into account the mutual connections and ranking orders among videos with similar durations. This approach may yield unstable results in the presence of sparse or uneven distributions of average watch times, as mentioned in Figure~\ref{fig:analysis}.
Zhan \etal \cite{kuaishou} propose to train the videos of different durations separately to remove the bias of video duration. 
\begin{table}[t]
  \centering
  \scriptsize
  \setlength{\tabcolsep}{6pt}
  \begin{tabular}{l|ccc|cc}
    \toprule
    & \multicolumn{3}{c|}{Content} &  \multicolumn{2}{c}{Metrics} \\
     & Video & Audio & Text &  Annotators number & Metric Sources \\ \midrule
    VQA datasets & \cmark & \xmark & \xmark & $\leq$ 40 & Labeling Scores \\
    Our datasets & \cmark & \cmark & \cmark & $\geq$ 2000 & Real User Interactions \\ 
    \bottomrule
  \end{tabular}
  \caption{We provide a detailed comparison with the VQA datasts. Our dataset contains multi-modal content to better measure the quality of videos. Moreover, our metrics are derived from thousands of real-world user interactions.}\label{tab:dataset1}
\end{table}
\section{SnapUGC Engagement Dataset}
\subsection{Pilot Study}
 To model the engagement levels of the videos, we initially explore the use of mainstream video quality assessment (VQA) methods, commonly used for evaluating video quality. We conducted assessments using state-of-the-art video quality assessment methods \cite{uvq, Wu_2023_ICCV} on a collection of real-world UGC short videos sourced from Snapchat Spotlight. These VQA methods were originally pre-trained on diverse VQA datasets \cite{uvq,patch-vq}. 
As shown in Wu \etal \cite{Wu_2018_beyond}, the average watch time can reflect the engagement levels of the videos with similar durations. Consequently, we conduct an evaluation aimed at evaluating the generalization capability of models trained on VQA datasets by calculating the correlation between Mean Opinion Score (MOS) and the engagement levels.
To mitigate the potential influence \cite{Wu_2018_beyond} of video duration, we categorize the real short videos into distinct groups based on their respective durations. Within each group of similar durations, we assessed the correlation between the average watch time and the predicted MOS scores for videos.
Our observation, as shown in Table~\ref{tab:pilot_study}, reveals a lack of correlation between the learned quality of pre-trained VQA methods and the engagement levels of the videos. This observation demonstrates that existing MOS scores provided by mainstream VQA datasets have difficulties in accurately reflecting the engagement levels.

\subsection{Dataset Collection}
While several previous datasets \cite{Wu_Speech2lip,Gupta_2022_CVPR, JiangHLLLL2019} are proposed for applications on short videos, they do not focus on video engagement analysis.
To precisely model the engagement levels of real UGC short videos, we first collect a large-scale short video dataset, named SnapUGC. Our dataset comprises 90,000 short videos, all of which were published on Snapchat Spotlight. For each video, we have curated corresponding aggregated engagement data derived from viewing statistics.
All short videos in our dataset have a duration ranging from 10 to 60 seconds. To mitigate sampling bias from small number of views, only short videos with view numbers exceeding 2000 are selected.
The dataset is notably diverse, encompassing a wide range of video types, including Family, Food \& Dining, Pets, Hobbies, Travel, Music Appreciation, Sports, etc. Several frames are shown in Figure~\ref{fig:sample}. We provide a comprehensive comparison with traditional VQA datasets in Table~\ref{tab:dataset1}.

\begin{figure*}[!t]
    \tiny
    \begin{center}
    \setlength{\tabcolsep}{0.5pt}
    \begin{tabular}{@{} c c @{}}
    \begin{tabular}{@{} c @{}}
               \includegraphics[width=0.04\linewidth]{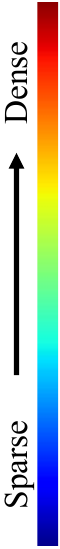} 
    \end{tabular} &
    \setlength{\tabcolsep}{0pt}
    \begin{tabular}{@{} c c c @{}}
    \includegraphics[width=.31\linewidth]{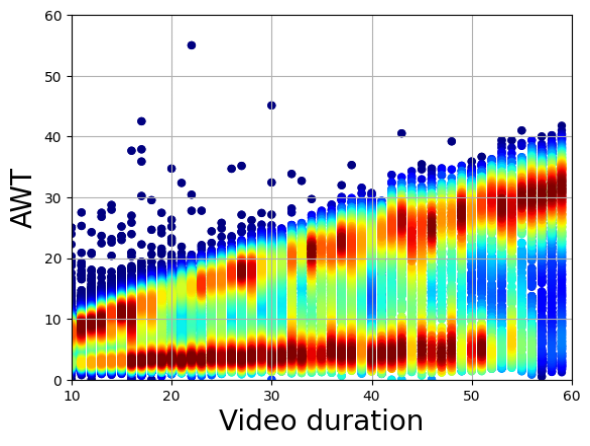} &
         \includegraphics[width=.31\linewidth]{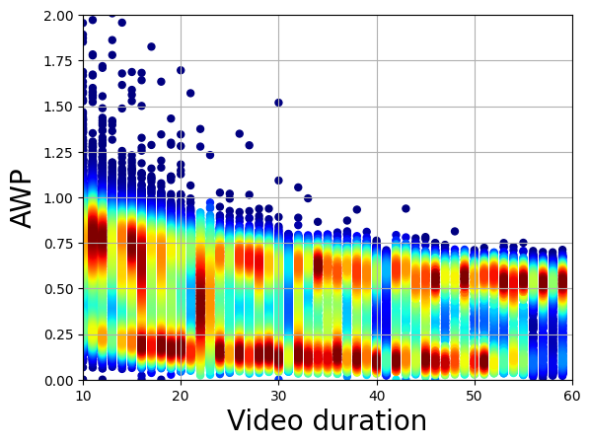} &
         \includegraphics[width=.31\linewidth]
         {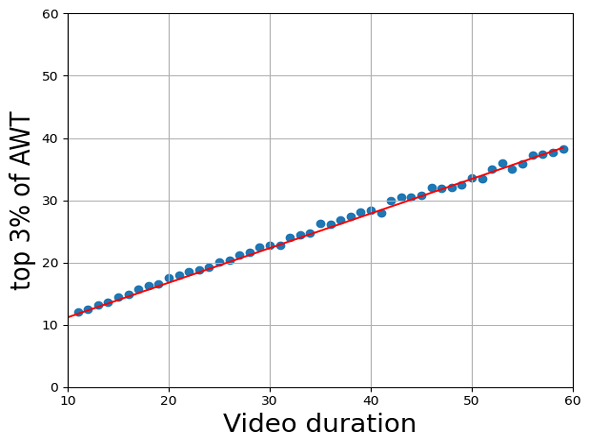} \\
         (a) Average watch time (AWT). & (b) Average watch percentage (AWP). & (c) Fitting top 3 \% of AWT. \\
         \includegraphics[width=.31\linewidth]
         {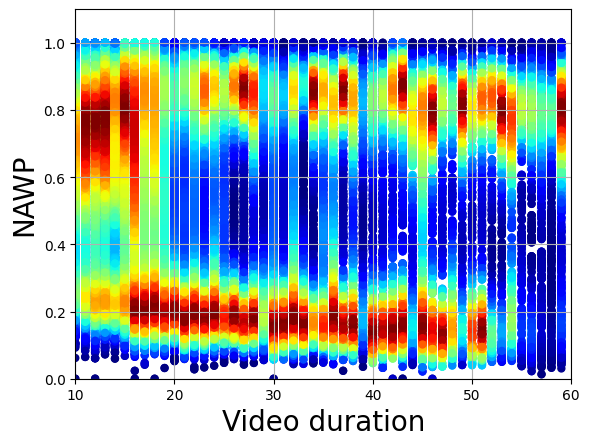} & 
         \includegraphics[width=.31\linewidth]
         {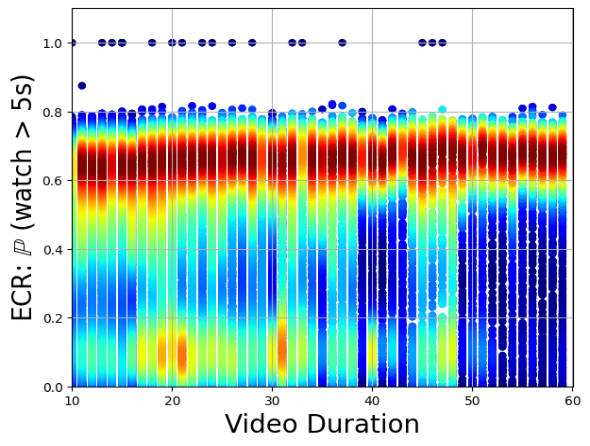} & 
     \includegraphics[width=.31\linewidth]{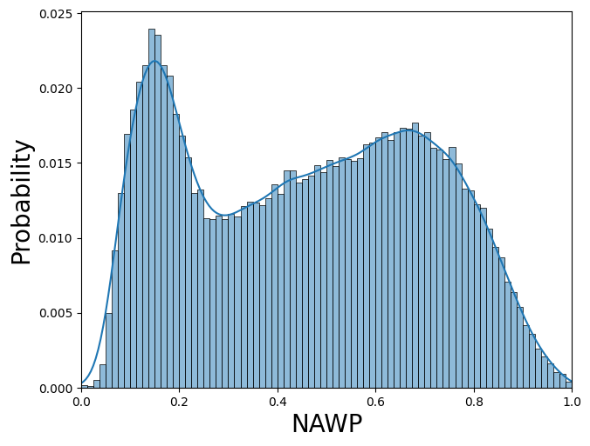} \\
        (d) Normalized average watch & (e) Engagement continuation rate & (f) NAWP follows a bimodal \\
        percentage (NAWP) & (ECR) $\mathbb{P}$ (watch \textgreater 5s). & distrbution. \\ 
        \includegraphics[width=.31\linewidth]
         {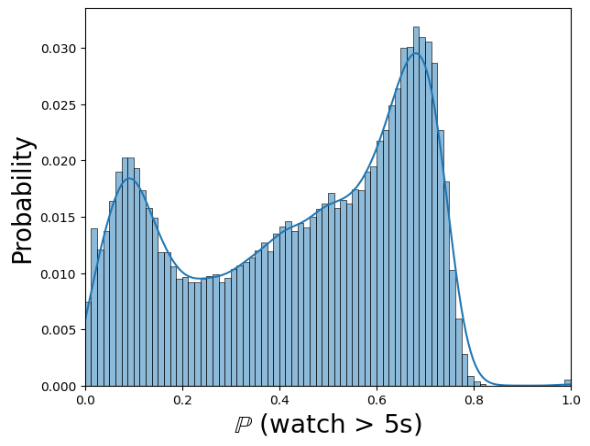} & 
         \includegraphics[width=.31\linewidth]
         {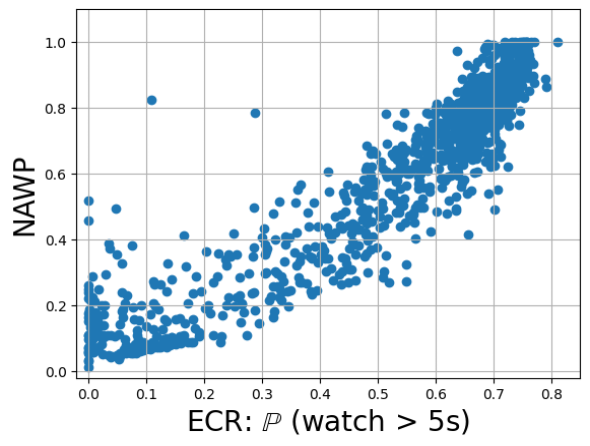} & 
     \includegraphics[width=.31\linewidth]{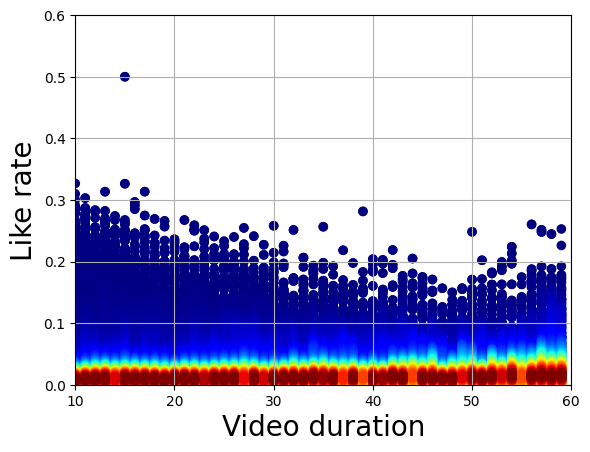} \\
     (g) ECR follows a bimodal & (h) Correlation between ECR & \multirow{2}{*}{(i) Like rate.} \\
        distribution. & and NAWP. &  \\
    \end{tabular}
    \end{tabular}
    \end{center}
    \caption{(a), (b), (e): The distributions of average watch time (AWT), average watch percentage (AWP) and engagement continuation rate (ECR), respectively. ECR, calculated as the probability of watch time exceeding 5 seconds: $\mathbb{P}$ (watch \textgreater~5s), is more duration-independent. 
    (c): We fit top 3\% of average watch times to derive a universal metric for videos of different durations. (d): Further normalization of the average time is achieved by fitting a line, resulting in the normalized average watch percentage (NAWP). 
    A color mapping is used to encode the distribution densities in (a), (b), (d) and (e). (f), (g): Distributions of NAWP and ECR. Both two metrics follow bimodal distribution, reflecting the unique property of user's swiftly skipping uninteresting videos or spend relative longer time on their interesting videos in short videos platforms. (h): The strong correlation between ECR and NAWP. (i): The distribution of like rate.}
  \label{fig:analysis}
\end{figure*}

\subsection{Engagement Metrics Analysis}
For short videos, there are three straightforward metrics to measure viewer engagement: view numbers, like rates, and average watch time. However, each metric has its drawbacks. View numbers can be heavily influenced by recommendation systems, leading to potential bias. Short videos created by well-known content creators may receive significantly higher view numbers compared to those of new creators. Like rates, although reflective of viewer interest, often yield extremely small and indistinguishable values across different videos, posing challenges for effective learning. A detailed study on like rates is shown in Figure~\ref{fig:analysis}(i) and supplementary. Average watch time (AWT), while common, faces limitations when comparing videos of different durations.
In this section, we first analyze the distribution and drawback of AWT, and then propose normalized average watch percentage (NAWP) as a novel engagement metric. Recognizing that users swiftly navigate through uninteresting content but persist in watching engaging videos, we introduce an additional metric: engagement continuation rate (ECR). Calculated for each video, this metric represents \emph{the proportion of viewers who watched the video for at least 5 seconds}. It serves as an indicator of a video's ability to captivate viewers at the beginning. Unlike Kim \etal \cite{MOOC} measuring entire videos' dropout probability, ECR focuses on 
he contents of first several seconds, which determines whether the users would continue to watch and substantially affects watch times. 
The experiment in Table~\ref{tab:joint} also demonstrates the effectiveness of ECR on help learning NAWP.

\noindent\textbf{Average watch time (AWT).}
We analyze average watch times (AWT) of various video durations $d$ in Figure~\ref{fig:analysis}(a). 
A similar metric, average watch percentage (AWP), is calculated as AWT divided by $d$, and its distribution with video duration is shown in Figure~\ref{fig:analysis}(b). When the AWT of a video surpasses its duration, AWP exceeds 1, signifying that the video is popular to be watched repeatedly.
Importantly, the distributions of AWT and AWP vary for different video durations, showing diverse user engagement patterns. Videos exhibit decreasing AWP as video duration increased, suggesting users' reduced likelihood of watching longer videos, potentially a result of declining attention spans.
Due to this duration-dependent behavior, comparing the popularity of short videos with different durations using AWT or AWP is challenging. For instance, a 30-second video with an AWT of 30 seconds and a 60-second video with the same watch time tend to have different engagement levels. Similarly, a 10-second video with an AWP of 1.0 and a 30-second video with an AWP of 1.0 may differ in engagement levels, because a shorter video is easier to be fully watched.

\noindent\textbf{Normalized average watch percentage (NAWP).} We introduce a straightforward metric called normalized average watch percentage (NAWP) to provide a generalized measure for videos with different durations. 
It is observed in Figure~\ref{fig:analysis}(a) that the largest values under different durations align with a linear trend. Based on the observation, we make the assumption that videos with top 3\% of highest AWT, regardless of their durations, are equally most popular, while videos with an average watch time of 0 seconds are deemed the least popular.
For example, a 40-second video with an AWT of 30 seconds and a 60-second video with an AWT of 40 seconds are regarded as equally most popular. Similarly, a 40-second video with an AWT of 0 seconds and a 60-second video with an AWT of 0 seconds are considered equally least popular.
The maximum average watch time $f_{\text{max}}(d)$ for most popular videos and minimum average watch time $f_{\text{min}}(d)$ for the least popular videos can be modeled by two linear functions:
\begin{equation}
   f_{\text{max}}(d) = 0.556 \times d + 5.64;~f_{\text{min}}(d) = 0.
\end{equation}
$f_{\text{max}}(d)$ is shown in Figure~\ref{fig:analysis}(c). The NAWP for any video of $d$ seconds, with average watch time $t$ is derived through normalization between $f_{\text{min}}(d)$ and $f_{\text{max}}(d)$:
\begin{equation}
    \text{NAWP}(\text{AWT},d) = \min\left(\frac{\text{AWT} - f_{\text{min}}(d)}{f_{\text{max}}(d)-f_{\text{min}}(d)},1\right). 
    \label{eq:clip}
\end{equation}
The relationship between the video duration and NAWP is depicted in Figure~\ref{fig:analysis}(d). The NAWP falls within the range of [0, 1] and NAWP of videos with top 3\% average watch time is set to be 1. The experiments in Table~\ref{tab:NAWP} shows that training with NAWP achieves much better performances than AWT or AWP.

\noindent\textbf{Engagement continuation rate (ECR).} As shown in Figure~\ref{fig:analysis}(e), engagement continuation rate (ECR), calculated as $\mathbb{P}$ (watch \textgreater 5s), demonstrates stable behavior across different video durations. The majority of values fall within the range of [0, 0.8]. The observation aligns with the metric's focus on frames within first 5 seconds. Furthermore, we observe a robust correlation of 0.926 between ECR and NAWP, as shown in Figure~\ref{fig:analysis}(h). Videos with higher probabilities of watch time surpassing 5 seconds tend to exhibit longer average watch times, illustrating a strong correlation between these two metrics. This finding offers valuable insights for designing the network structure and joint training strategy, to be shown in Section~\ref{Sec:4_3} and Table~\ref{tab:joint}.

\noindent\textbf{Bimodal distributions.} It is observed in Figure~\ref{fig:analysis}(f) and (g), that distributions of NAWP and ECR exhibit a bimodal pattern. Compared with the single peak distribution of MOS scores \cite{MD-VQA,icme2021}, this bimodal distribution is \textbf{unique} to our dataset.
This behavior exists due to the common UI designs that encourages ``swiping'' to skip boring videos in short video platforms. Users usually quickly skip through uninteresting videos, whereas they tend to dedicate relatively longer time to engaging with videos they find interesting. Consequently, it results in two separate peaks in the distributions.

\noindent\textbf{Generalizability of NAWP.}  
While NAWP is designed based on the linearity observation on our SnapUGC dataset, 
It is obversed in supplementary that \emph{the linear approximation} can generalize to average watch time of \emph{Kuaishou \cite{kuaishou} and Youtube \cite{Wu_2018_beyond} datasets} for videos with short durations ($\leq$ 60s), which are exactly the domain of most short videos, explored in this paper.

\section{Methods}
In this section, we begin by presenting the natural bias of recommendation systems and formulate the engagement prediction. Then we conduct an in-depth exploration of the multi-modal features that aid engagement prediction in Section~\ref{Sec:4_2}. In Section~\ref{Sec:4_3}, we provide details about our network, and in Section~\ref{Sec:4_4}, we outline the evaluation criteria.
\begin{figure}[!t]
  \centering
    \includegraphics[width=0.9\linewidth]{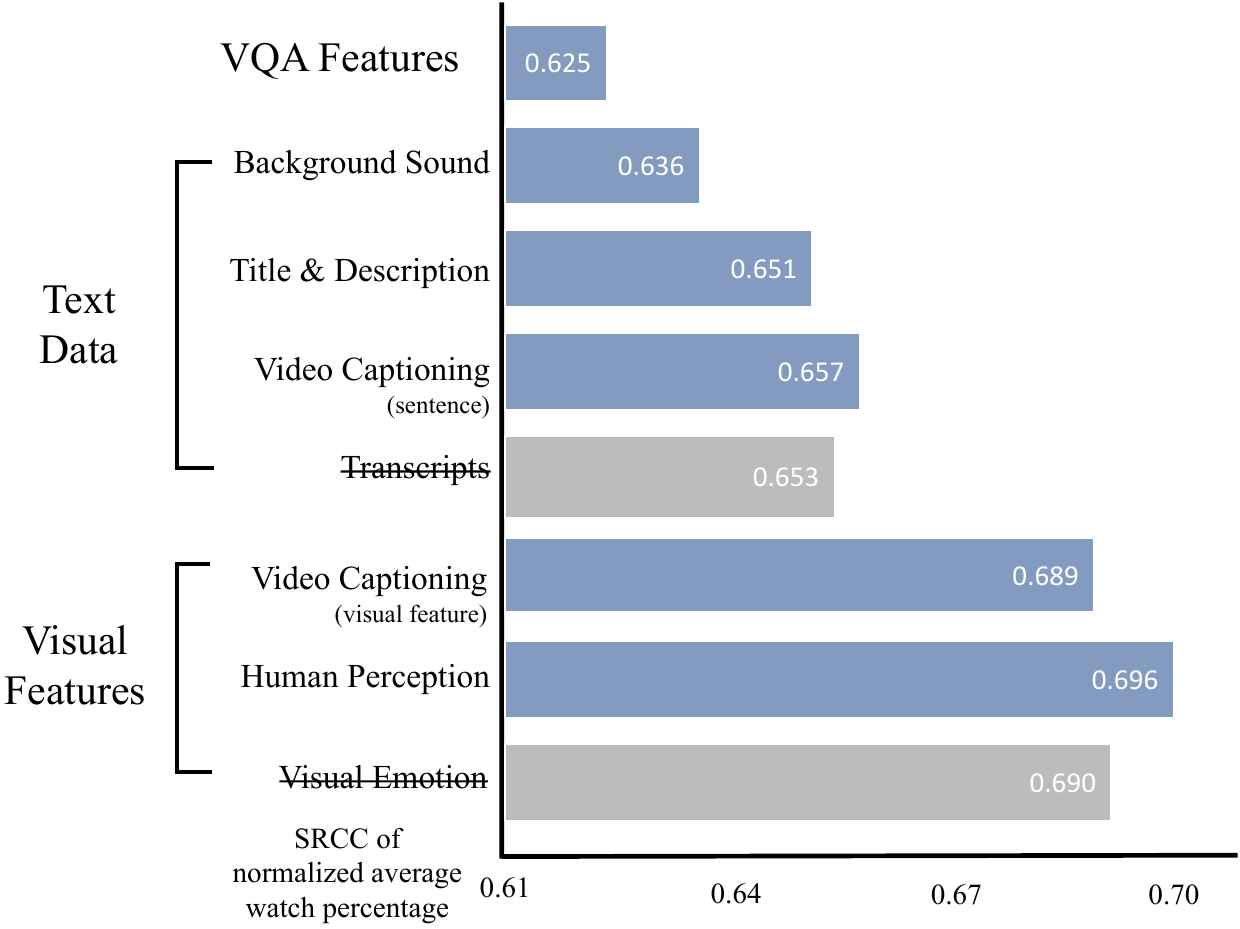}
    \caption{The effectiveness of comprehensive multi-modal features to enhance engagement prediction. The blue bars represent incrementally incorporating new features to achieve improved SRCC, while a gray bar indicates that the modification was not adopted. These multi-modal features incorporated into our network leads to increasingly better performance than previous VQA features.} 
    \label{fig:investigate}
\end{figure}

\subsection{Problem Formulation}
Notably, the normalized average watch percentage (NAWP) and engagement continuation rate (ECR) are contingent upon the recommendation system, denoted as $\mathbf{R}$. Recommendation systems often employ machine learning classifiers \cite{Gupta_2022_CVPR,resnet3d} to categorize short videos and analyze user preferences based on their historical engagements with various video types. These systems balance exploitation (recommending familiar contents and familiar creators) and exploration (introducing new contents and creators) to users. Consequently, the preference distribution for a given short video may vary depending on the exploitation strategy employed by different recommendation systems. The engagement metrics are biased due to the preference distribution provided by the recommendation system $\mathbf{R}$.
Therefore, we formulate engagement prediction as a realistic conditional problem. For a given short video $v$ and the recommendation system $\mathbf{R}$, our network $G$ predicts the normalized average watch percentage $\widehat{\text{NAWP}}$ and the engagement continuation rate $\widehat{\text{ECR}}$ as follows:
\begin{equation}
    (\widehat{\text{NAWP}},~\widehat{\text{ECR}}) = G(v~|~\mathbf{R}). 
\end{equation}
We only focus on aggregated metric in this work as individual user's metric is subject to legal and privacy concerns.

\subsection{Comprehensive Features for Engagement Prediction}
\label{Sec:4_2}
To precisely model the engagement levels of short videos, we investigate a comprehensive set of multi-modal features. The evaluation of various features is conducted using the Spearman Rank Correlation Coefficient (SRCC) of the normalized average watch percentage (NAWP). We utilize T5 \cite{2020t5} as the text encoder to encode the text data.
In Figure~\ref{fig:investigate}, we show the procedure and the incremental performance achieved by gradually incorporating each feature.
In particular, our exploration focuses on the following aspects:

\begin{itemize}
    \item \textbf{VQA features.} Building on established video quality assessment methods UVQ \cite{uvq} and MD-VQA \cite{MD-VQA}, we extract per-frame semantic features \cite{efficientnet_v2} per-frame distortion features \cite{uvq}, and action recognition features \cite{resnet3d} for video clips. These features collectively offer a fundamental assessment of both content and objective quality. This baseline gives a correlation of 0.625.
    \item \textbf{Background sound.} Creators usually incorporate background music in short videos to enhance the atmosphere and attract viewers. We employ YAMNet \cite{tensorflow2015-whitepaper}, a 521-class audio event classification model, to discern various types of background music. The top 5 classification results, presented as text, are then utilized as an additional network input to augment the modeling of video engagement. This improves the performance from 0.625 to 0.636.
    \item \textbf{Title and descriptions} are usually provided along with the short videos by the creators, which can emphasize key content and provide additional context information, enhancing the overall understanding of the videos. Incorporating the title and description leads to an increase from 0.636 to 0.651.
    \item \textbf{Video captioning.}
    Video captioning provides fine-grained understanding of the short videos. Leveraging mid-layer features and captions generated by mPLUG-2 \cite{Xu2023mPLUG2AM} as complementary features enhances engagement predictions. The captions would also provide new insights for interpreting video popularity. The inclusion of captions increases the performance slightly to 0.657. Adding intermediate features as additional input visual features brings a significant improvements from 0.657 to 0.689.
    \item \textbf{Transcripts.} Ideally, transcripts would facilitate a better understanding of video content. However, our findings indicate that adding transcripts does not yield improvements. This observation can be attributed to the fact that only 30\% of short videos include effective transcripts. Additionally, viewers often decide whether to continue watching based on the initial seconds, during which they only catch a small amount of the spoken content.
    \item \textbf{Human asethetic preference.} 
    While the semantic \cite{efficientnet_v2} and action features \cite{resnet3d} described above contain semantic information,
    they may not directly capture human reactions and feelings when watching videos. In response, Wu \etal \cite{Wu_2023_ICCV} proposed a mean aesthetic option score to measure human quality opinions solely from an aesthetic perspective. Leveraging human aesthetic preferences may contribute to modeling the popularity of short videos. 
    Therefore, we integrate the aesthetic features extracted from pretrained models in \cite{Wu_2023_ICCV}, resulting in an increase from 0.689 to 0.696.
    \item \textbf{Visual emotion.} Creators often convey emotions through short videos, and these emotions can be reflected in the visual sentiment captured in individual frames. 
    To evaluate the potential benefits of  emotion information, we employ WSCNet \cite{Yang_2018_CVPR, She_2019_TMM}, trained on the WEBEmo dataset \cite{panda2018contemplating}  to obtain intermediate features. The observed change from 0.696 to 0.690 suggests a limited correlation between visual sentiment and engagement levels.
\end{itemize}
\begin{figure}[!t]
  \centering
    \includegraphics[width=1.0\linewidth]{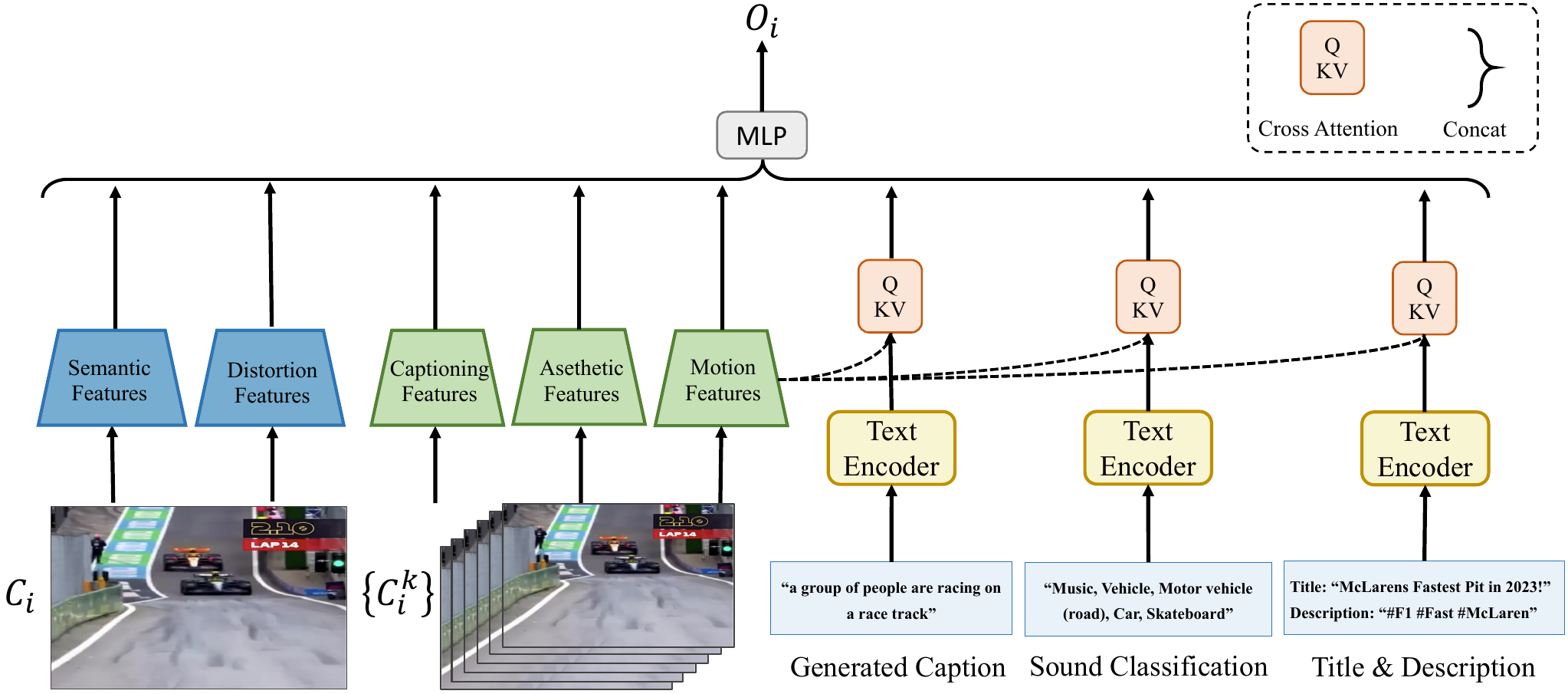} 
    \caption{The overview of multi-modal feature extractions. The learnable Multilayer Perceptron (MLP) to process extracted features is omitted for simplicity.}
    \label{fig:investigate1}
\end{figure}

\subsection{Network Details}
\label{Sec:4_3}
Following MD-VQA \cite{MD-VQA}, we split the video into several clips for efficient feature extraction. 
Given a video with frame count $M$ and frame rate $r$, we create $\frac{M}{L}$ clips $\{C_i\}_{i=1}^{M/L}$ with each clip $C_i$ containing $L$ frames $\{C_i^k\}_{k=1}^{L}$. Our network takes visual features and text data as inputs. The feature extraction is shown in Figure~\ref{fig:investigate1}.
For each clip $C_i$, we extract semantic and distortion features for each of the $L$ frames ${C_i^{k}}$, while the entire clip $C_i$ is used for action feature extraction, asethetic feature extraction and video captioning feature extraction. The text data, which include background sound classification, title, descriptions, and generated captions, are shared among all the clips. We process the visual features with learnable Multi-Layer Perceptrons (MLP) and employ cross-attention to merge visual action features with text data. Then the multi-modal features are fused by 8 MLP layers to obtain the fused features $\{O_i\}_{i=1}^{M/L}$.
Subsequently, we utilize a 8-layer self-attention architecture to combine the fused features $\{O_i\}_{i=1}^{M/L}$ of all the clips to obtain temporal aggregated features $\{H_i\}_{i=1}^{M/L}$. 
Finally, our network utilizes 2 MLP layers $F_{\text{out}}^1, F_{\text{out}}^2$ to jointly predict $\widehat{\text{NAWP}}$ and $\widehat{\text{ECR}}$:
\begin{equation}
    \widehat{\text{NAWP}} = \frac{L}{M} \sum_{i=1}^{M/L} F_{\text{out}}^1(H_i); \widehat{\text{ECR}} = \frac{L}{5r} \sum_{i=1}^{5r/L} F_{\text{out}}^2(H_i),
\end{equation}
where $\widehat{\text{ECR}}$ is derived from frames within first 5 seconds. The joint training loss $L$ for $\widehat{\text{NAWP}}$ and $\widehat{\text{ECR}}$ is derived:
\begin{equation}
    L = || \text{NAWP} - \widehat{\text{NAWP}} ||_2 + ||\text{ECR}- \widehat{\text{ECR}}||_2.
    \label{eq:loss}
\end{equation}

It is observed in our experiments (Table~\ref{tab:joint}) that training these two highly correlated metrics jointly leads to enhanced overall performance.

\subsection{Evaluation Criteria}
\label{Sec:4_4}
We evaluate our method using common criteria in Video Quality Assessment (VQA) research, including Spearman Rank Correlation Coefficient (SRCC), Pearson Linear Correlation Coefficient (PLCC) and Root Mean Square Error (RMSE) for both NAWP and ECR. 
Drawing insights from the observations in Figure~\ref{fig:analysis}(f) and (g), we empirically note that only 10\% to 20\% of uploaded short videos, centered around the second peak of the bimodal distributions, emerge as popular and are prioritized by the recommendation system. Therefore, we consider the top $K$\% of $N$ test videos with the highest $\text{NAWP}$. For these selected videos $\{v_j\}_{j=0}^{K\times N/100}$, we calculate RMSE of top 10\% of NAWP as follows:
\begin{equation}
    \text{RMSE}_{\text{top10\%}} = \sqrt{\frac{(\sum_{j=0}^{K\times N/100} (\widehat{\text{NAWP}_j} - \text{NAWP}_j)^2)} {K\times N / 100}},
\end{equation}
which is similar for the RMSE of top $K$\% of ECR. (K=10 in our evaluation.)
\begin{table*}[t]
  \centering
  \scriptsize
  \setlength{\tabcolsep}{2pt}
  \begin{tabular}{l|ccc|ccc|c|c}
    \toprule
    \multirow{2}{*}{Method} & \multicolumn{3}{c|}{NAWP} & \multicolumn{3}{c|}{ECR} & NAWP &  ECR \\ 
     & SRCC$\uparrow$ & PLCC$\uparrow$ & RMSE$\downarrow$  & SRCC$\uparrow$ & PLCC$\uparrow$ & RMSE$\downarrow$ & RMSE$_{\text{top 10\%}}\downarrow$ & RMSE$_{\text{top 10\%}}\downarrow$ \\ \midrule
    VSFA \cite{vsfa} & 0.609 & 0.615 & 0.192 & 0.576 & 0.591 & 0.197 & 0.199 & 0.174 \\ 
    PVQ \cite{patch-vq} &  0.590 & 0.607 & 0.197 & 0.587 & 0.602 & 0.194 & 0.189 & 0.170 \\ 
    MD-VQA \cite{MD-VQA} & 0.606 & 0.614 & 0.193 & 0.592 & 0.608 & 0.191 & 0.187 & 0.166 \\ 
    FastVQA \cite{wu2022fastquality} & 0.587 & 0.590 & 0.218 & 0.581 & 0.585 & 0.223 & 0.232 & 0.201 \\ 
    DOVER \cite{Wu_2023_ICCV} & 0.635 & 0.636 & 0.206 & 0.619 & 0.622 & 0.203 &  0.216 & 0.189 \\ \midrule
    Ours-VQA & 0.625 & 0.632 & 0.188 & 0.605 & 0.620 & 0.189 & 0.191 & 0.171\\
    Ours & \textbf{0.696} & \textbf{0.701} & \textbf{0.172} & \textbf{0.675} & \textbf{0.688} & \textbf{0.174} & \textbf{0.181} & \textbf{0.152}  \\ 
    \bottomrule
  \end{tabular}
  \caption{Experimental performances of NAWP and ECR on the proposed engagement prediction dataset. ``Ours-VQA'' denotes merely utilizing VQA features (per-frame semantic features, per-frame distortion features and per-clip action recognition features).}
  \label{tab:results}
\end{table*}

\section{Experiments}

\subsection{Implementation Details}
For the SnapUGC dataset, we adhere to the common practice and spilt the dataset with an 90\%$\sim$10\% train-test ratio. Our network $G$ takes extracted features as input and regresses $\widehat{\text{NAWP}}$ and $\widehat{\text{ECR}}$. All feature extraction networks are pre-trained separately. We follow UVQ \cite{uvq} to train a distortion recognition network on KADIS-700K and KADID-10K \cite{kadid}. The per-frame semantic features are extracted by EfficientNet \cite{efficientnet_v2}, pre-trained on ImageNet \cite{imagenet}. The per-clip action recognition features are extracted by ResNet-3D \cite{resnet3d}, pre-trained on Kinetics-400 \cite{k400data}. The video caption and mid-layer features are extracted from the pre-trained video captioning model mPLUG-2 \cite{Xu2023mPLUG2AM}. Human aesthetic features are extracted by the pre-trained model in DOVER \cite{Wu_2023_ICCV}. We utilize T5 \cite{2020t5} as a text encoder to encode the text data, including generated captions, sound classification results, titles, and descriptions. The network is trained with a batch size of 8 for 70,000 iterations. We use the Adam optimizer~\cite{adam} and the learning rate is decreased from $1\times10^{-4}$ to $1\times 10^{-7}$ according to the cosine annealing strategy~\cite{cosine_annealing}. The parameter $L$ is set to be 16. We optimize $\widehat{\text{NAWP}}$ and $\widehat{\text{ECR}}$ 
jointly following Eq (\ref{eq:loss}). More details are provided in supplementary materials.
\subsection{Engagement Results}
To evaluate the performance of the proposed framework, we select popular quality assessment methods for comparisons, including VSFA \cite{vsfa}, PVQ \cite{patch-vq}, MD-VQA \cite{MD-VQA}, FastVQA \cite{wu2022fastquality}, and DOVER \cite{Wu_2023_ICCV}. \textit{Our network and these VQA methods are trained} on the proposed engagement dataset to learn the normalized average watch percentage (NAWP) and engagement continuation rate (ECR) jointly. 
As conventional distortion features in MD-VQA \cite{MD-VQA} are not available, we substitute them with distortion networks from UVQ \cite{uvq}. We enhance the models by adding an additional final layer of VSFA \cite{vsfa}, PVQ \cite{patch-vq}, and MD-VQA \cite{MD-VQA} to make them adaptive for joint training with two metrics.
Due to the frames sampling in FastVQA \cite{wu2022fastquality} and DOVER \cite{Wu_2023_ICCV}, we train two separate models to predict NAWP and ECR. The sampling range for NAWP is set to frames of whole videos, while the sampling range for ECR is set to frames within the first 5 seconds. All VQA models are trained with the default parameters defined by their respective authors.
\begin{table}[t]
  \centering
  \scriptsize
  \setlength{\tabcolsep}{11pt}
  \begin{tabular}{cc|c}
    \toprule
    Learning metrics & Duration as input & Average SRCC \\ \midrule
   AWP & \xmark & 0.665~~~~~~ \\ 
    AWP & \cmark & 0.681 (\textbf{$\uparrow$}) \\ \midrule
    AWT & \xmark & 0.668~~~~~~ \\ 
    AWT & \cmark & 0.683 (\textbf{$\uparrow$}) \\ \midrule
    NAWP & \xmark & \textbf{0.696}~~~~~ \\ 
    NAWP & \cmark & 0.689 (\textbf{$\downarrow$}) \\ 
    \bottomrule
  \end{tabular}
  \caption{We compare proposed NAWP with average watch percentage (AWP) and average watch time (AWT). ``Duration as input'' means adding the video duration as a network input. We divide the videos to different groups according to their video durations and average the SRCC of different groups to obtain ``Average SRCC''.}
  \label{tab:NAWP}
  \setlength{\tabcolsep}{22pt}
  \begin{tabular}{l|cc}
    \toprule
    Training setting & NAWP & ECR \\ \midrule
    Separate training & 0.662 & 0.681 \\
    Joint training & \textbf{0.675} & \textbf{0.696} \\ 
    \bottomrule
  \end{tabular}
  \caption{Ablation of joint training normalized average watch percentage (NAWP) and engagement continuation rate (ECR).}
  \label{tab:joint}
\end{table}

The experimental performance on the proposed dataset is shown in Table~\ref{tab:results}. 
``Ours-VQA'' denotes the model merely incorporating VQA features (per-frame semantic features, per-frame distortion features and per-clip action recognition features). The difference between ``Ours-VQA'' and MD-VQA \cite{MD-VQA} lies in the utilization of self-attention layers \cite{attention}, resulting in a 0.17 improvement in SRCC for NAWP.
Although DOVER \cite{Wu_2023_ICCV} outperforms `Ours-VQA' on SRCC, it exhibits significantly poorer results on RMSE. ``Ours-VQA'' achieves balanced performance across SRCC, PLCC, and RMSE.
Benefiting from the integration of complementary multi-modal features, our method outperforms state-of-the-art VQA models by a clear margin.

\subsection{Ablation Study}

\noindent\textbf{Normalized average watch percentage.} To evaluate the performance of normalized average watch percentage (NAWP), we train two models with average watch time (AWT) and average watch percentage (AWP). Since both AWT and AWP are duration-dependent metrics, calculating the correlation among videos of different durations is not feasible. 
Therefore, we categorize videos into groups based on their durations and compute SRCC for average watch percentage within each group.
The average SRCC across these groups served as the evaluation metric in this ablation study. Table~\ref{tab:NAWP} illustrates that the proposed NAWP outperforms AWT and AWP by a significant margin.
We also explore incorporating video duration as a network input, as Wu \etal \cite{Wu_2018_beyond} do. Although incorporating video duration as input leads to improvements in the learning performance associated with AWT and AWP, their performances are still worse than learning the proposed NAWP. Given that NAWP is more duration-independent, the incorporation of video duration as a network input cannot yield better improvements and can lead to potential confusion and overfitting. Furthermore, the models trained with AWT and AWP are unsuitable for comparing two videos with different durations, as detailed in supplementary materials.

\noindent\textbf{Jointly training with two metrics.} 
We conducted an experiment between joint training of two metrics and separate training of each metric. As illustrated in Table~\ref{tab:joint}, joint training significantly enhances the performance of both metrics. The boost performance indicates that the strong correlation between the two metrics contributes to the ability of joint training to achieve higher performances.

\section{Conclusion}
In this paper, we first reveal the limitation of using mean opinion scores from previous video quality datasets to model popularity. To overcome this, we curate a large-scale dataset of real-world short videos and conduct a detailed analysis of engagement metrics and their correlations. We further investigate comprehensive multi-modal features to enhances the model's performance. The resultant model achieves state-of-the-art performance in predicting engagement for short videos.


\clearpage  

%
%
\bibliographystyle{splncs04}
\bibliography{main}
\end{document}